\documentclass[sigconf,natbib=true,anonymous=false]{acmart}

\usepackage{algorithm}
\usepackage{algorithmic}
\usepackage{dblfloatfix}
\usepackage{capt-of}
\usepackage{paralist}
\usepackage{url}
\usepackage{xspace}
\usepackage{amsfonts}
\usepackage{amsmath}
\usepackage{booktabs}
\usepackage{multirow}
\usepackage{tabularx}
\usepackage{multicol}
\usepackage{subfigure}
\usepackage{lipsum}
\usepackage{makecell}
\usepackage{adjustbox}

\usepackage{xcolor}
\usepackage{latexsym}
\usepackage{pifont}
\usepackage{mdframed}
\AtBeginDocument{%
  \providecommand\BibTeX{{%
    \normalfont B\kern-0.5em{\scshape i\kern-0.25em b}\kern-0.8em\TeX}}}

\setcopyright{iw3c2w3g}

\acmDOI{https://arxiv.org/abs/2307.11278} 
\acmConference[arxiv]{arxiv}{March 2024}{arxiv} 
\acmYear{2024}
\acmBooktitle{arxiv} 

\begin{document}

\title{Generator-Retriever-Generator Approach for Open-Domain Question Answering}


\author{Abdelrahman Abdallah}
\email{Abdelrahman.Abdallah@uibk.ac.at}
\affiliation{%
  \institution{University of Innsbruck}
  \streetaddress{Innrain 52}
  \city{Innsbruck}
  \country{Austria}
}

\author{Adam Jatowt}
\email{adam.jatowt@uibk.ac.at}
\affiliation{%
  \institution{University of Innsbruck}
  \streetaddress{Innrain 52}
  \city{Innsbruck}
  \country{Austria}
}

\renewcommand{\shortauthors}{}

\begin{abstract}
Open-domain question answering (QA) tasks usually require the retrieval of relevant information from a large corpus to generate accurate answers. We propose a novel approach called Generator-Retriever-Generator (GRG) that combines document retrieval techniques with a large language model (LLM), by first prompting the model to generate contextual documents based on a given question. In parallel, a dual-encoder network retrieves documents that are relevant to the question from an external corpus. The generated and retrieved documents are then passed to the second LLM, which generates the final answer. By combining document retrieval and LLM generation, our approach addresses the challenges of open-domain QA, such as generating informative and contextually relevant answers. GRG outperforms the state-of-the-art generate-then-read and retrieve-then-read pipelines (GENREAD and RFiD) improving their performance by at least by +5.2, +4.2, and +1.6 on TriviaQA, NQ, and WebQ datasets, respectively.  We provide code, datasets, and checkpoints\footnote{https://github.com/abdoelsayed2016/GRG}.
\end{abstract}


\maketitle
\section{Introduction}

\begin{figure*}[t]
  \centering
  \includegraphics[width=1.7\columnwidth]{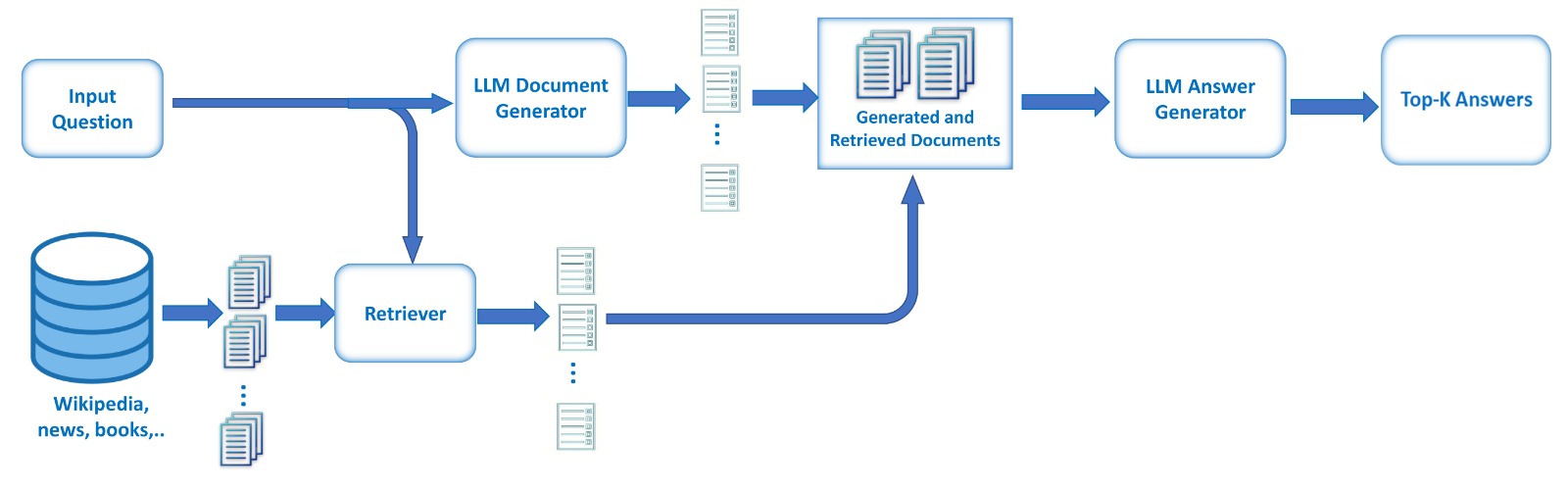}
  \caption{Simplified diagram illustrating the idea behind the Generator-Retriever-Generator approach.}
  \label{fig:figure_label1}
\end{figure*}

Open-domain question answering (QA) tasks pose significant challenges since they require access to large document collections or repositories of domain-specific knowledge. Existing methods for QA~\citep{karpukhin2020dense,izacard2020leveraging} often rely on a retrieve-then-read pipeline, where relevant contextual documents are retrieved from external sources like Wikipedia, and the answer prediction is conditioned on the retrieved documents and the question. These methods suffer however from several drawbacks. 
Firstly, the retrieved documents are often chunked and of fixed size, which can result in the inclusion of noisy and irrelevant information. The fixed-size document chunks may not adequately capture the context necessary for finding accurate answers~\citep{yu2022generate}. Consequently, the presence of irrelevant information can lead to noise in the retrieved documents, negatively impacting the quality and relevance of the generated answers. Secondly, the representations of questions and documents in current approaches are typically obtained independently~\citep{oguz2020unik}. This independent processing fails to capture the intricate interactions and dependencies between the question and the documents. As a result, the model's understanding of the question and its ability to extract relevant information from the retrieved documents may be limited. The shallow interaction between questions and documents hinders the model's capability to fully exploit the contextual cues present in the data, thereby limiting its answer generation accuracy.
The limitations on retriever model parameters and embedding sizes, imposed by the need to efficiently handle large corpora, restrict the model's capacity to fully leverage large language models' parametric knowledge and its deduction capabilities. Consequently, the retriever models may struggle to capture the rich semantic and contextual information necessary for accurate answer generation~\citep{levine2022standing}. 

On the other hand, open-domain QA often involves training a language model to generate answers for a given question without access to accompanying documents containing the answer~\citep{zhu2021retrieving}. One promising approach in open-domain QA is to augment the language model with an external knowledge source, such as Wikipedia, referred to as evidence documents~\citep{izacard2020leveraging}. This approach comprises two core components: an information retrieval system (the retriever) to identify relevant text snippets from the knowledge source and another system (the reader) to generate answers based on the retrieved documents and the question.

This paper proposes a novel approach called generator-retriever-generator (GRG) for open-domain question answering. Our method combines document retrieval techniques with large language models to address the challenges of generating informative and contextually relevant answers. We leverage the power of a large language model such as GPT3 and InstructGPT~\citep{brown2020language,ouyang2022training} to generate contextual documents based on a given question while simultaneously employing a dense passage retrieval system~\citep{singh2021end,karpukhin2020dense} to retrieve relevant documents from external sources. A second large language model then processes the generated and retrieved documents to produce the final answer. By integrating document retrieval and large language model generation, the proposed GRG approach aims to improve the accuracy of open-domain question answering. Fig. \ref{fig:figure_label1} shows the high-level architecture of the GRG approach.

Our contributions can be summarized as follows: 

\begin{enumerate}
    \item GRG Approach: We introduced the GRG approach that combines document generation and retrieval to improve answer generation in open-domain QA.
    \item Document Generation \& Retrieval Methods: We developed a method using InstructGPT for generating contextually rich documents. We also proposed the Vector Index Retriever for efficient retrieval of relevant documents.
    \item Effectiveness of GRG: We validated the effectiveness of our GRG approach through extensive experiments and analyses on three open-domain QA datasets.
\end{enumerate}   

\begin{table}[h!]
\centering
\caption{Advantages and Disadvantages of Question Answering Approaches}
\label{tab:advantages_disadvantages}
\resizebox{\columnwidth}{!}{%
\begin{tabular}{p{0.25\columnwidth}|p{0.35\columnwidth}|p{0.35\columnwidth}}
\hline
\textbf{Approach} & \textbf{Advantages} & \textbf{Disadvantages} \\
\hline
Retriever-Reader~\citep{karpukhin2020dense} & 
- Accesses extensive external knowledge.\newline
- Provides in-depth, context-based decision-making.
&
- Risk of overlooking relevant information.\newline
- Dependent on the accuracy and coverage of the retrieval process.
\\
\hline
Generator-Reader~\citep{yu2022generate} & 
- Capable of generating novel, contextually relevant answers.\newline
- Reduced dependence on pre-existing documents.
&
- High computational requirements.\newline
- Possible issues with the reliability and accuracy of generated answers.
\\
\hline
Retriever-Generator~\citep{izacard2020leveraging,singh2021end} & 
- Merges accurate retrieval with creative generation.\newline
- Enhances recall by supplementing existing content.
&
- Increased computational complexity.\newline
- Balancing quality and diversity of answers can be challenging.
\\
\hline
Retriever-Only~\cite{lee2020learning} & 
- Directly utilizes a broad range of existing documents.\newline
- Provides well-grounded, context-based responses.
&
- May miss crucial documents.\newline
- Limited flexibility in handling complex queries.
\\
\hline
Generator-Retriever-Generator & 
- Ensures high relevance and accuracy.\newline
- Adaptable to a wide range of queries.
&
- Significant computational demands.\newline
- Balancing diverse and high-quality answers can be difficult.
\\
\hline
\end{tabular}
}
\end{table}

\section{Related Work}

We describe in this section related works that fall into 4 known open-domain QA architectures: \textit{Retriever-Reader}, \textit{Generator-Retriever}, \textit{Generator-Reader}, and \textit{Retriever-only}.
\subsection{Retriever Reader}
The Retriever-Reader approach is based on the idea of combining information retrieval (retriever) and machine reading comprehension (reader) techniques. Previous work in this area includes the use of document retrieval techniques such as TF-IDF, BM25, or neural ranking models \citep{rosa2021yes,article_tiidf} to select relevant documents from a large corpus.  mNotable works include the  Stanford Question Answering Dataset (SQuAD) and subsequent advancements in retriever-reader architectures like DrQA and BiDAF~\citep{seo2018bidirectional}. Dense Passage Retrieval (DPR) \citep{karpukhin2020dense} focuses on dense representations for passage retrieval, utilizing a dual-encoder architecture to retrieve passages and a reader model to extract the answer. T5-RC \citep{raffel2020exploring}, a variant of the T5 model, follows the Retriever-Reader approach by retrieving relevant passages and applying T5 as a reader for answer extraction. 
\subsection{Retriever Generator}
The Retriever-Generator~\citep{izacard2020leveraging,singh2021end} approach aims to leverage both generative modeling and retrieval techniques. Previous work~\citep{zhu2021retrieving} in this direction has explored methods for retrieving supporting passages using sparse or dense representations. The retrieved passages are then used as input into a sequence-to-sequence model, such as a transformer-based architecture, which generates the answer to the question. This approach has shown improved performance on benchmark datasets like TriviaQA~\cite{joshi2017triviaqa} and NaturalQuestions~\cite{kwiatkowski2019natural}. 
\subsection{Generator Reader}
The Generator-Reader approach~\cite{yu2022generate} focuses on generating contextual documents based on a question and then using a reader model to extract the answer from the generated context. The approach involves training large language models, such as Generative Pre-trained Transformer (GPT)~\cite{radford2019language}, to generate coherent and relevant documents given a prompt. The generated documents are then processed by a reader component, which can be a reading comprehension model, to extract the answer.
On the other hand, the DocGen approach, introduced in~\cite{askari-etal-2023-expand}, focuses on generating synthetic documents from queries. The DocGen pipeline involves expanding and highlighting the original query before generating a synthetic document likely to be relevant to the query. To enhance the relevance between generated synthetic documents and their corresponding queries, the authors propose DocGen-RL. This method treats the estimated relevance of the document as a reward and uses reinforcement learning (RL) to optimize the DocGen pipeline.
\begin{figure}[t!]
  \centering
  \includegraphics[width=0.49\textwidth]{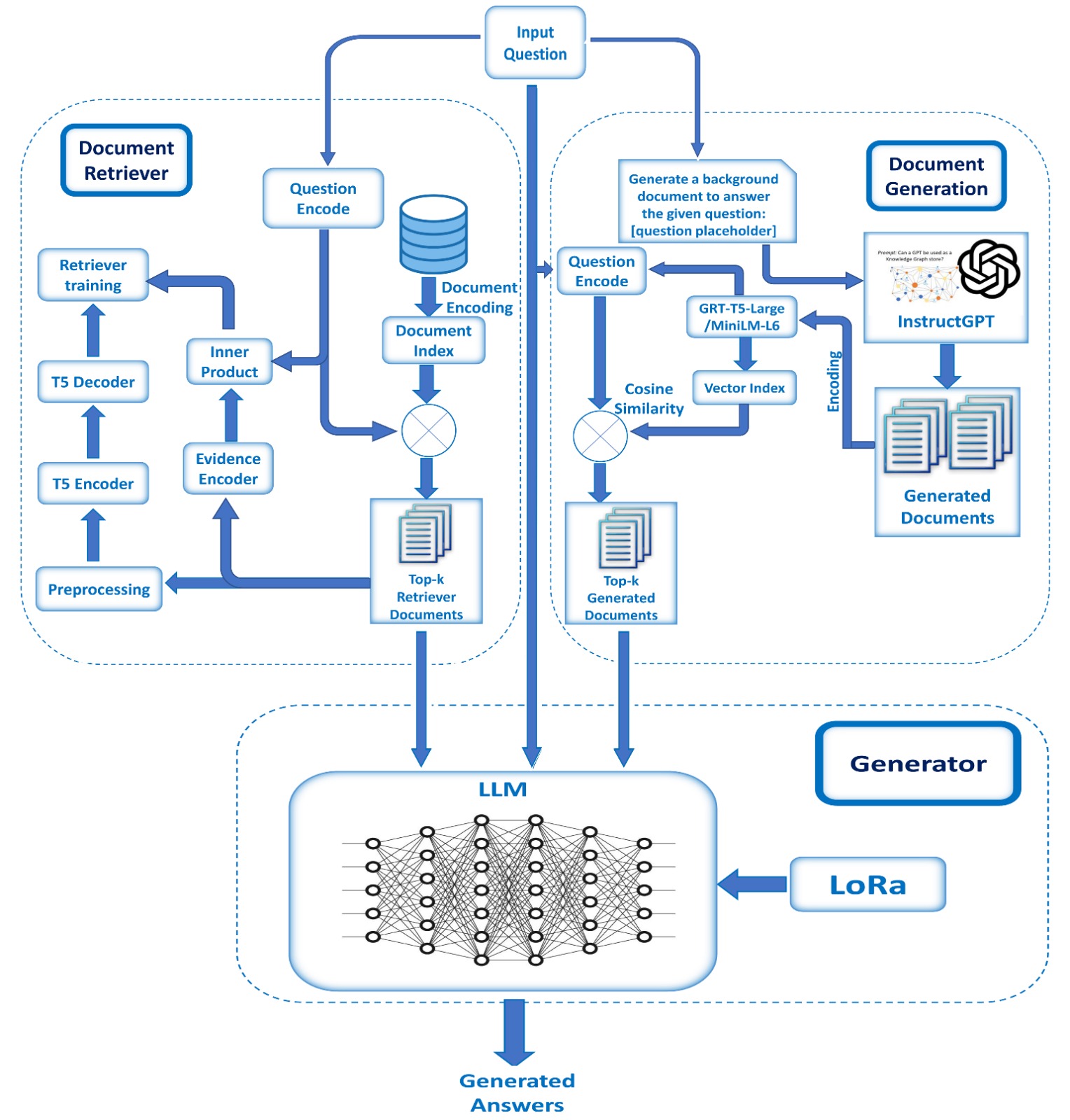}
  \caption{Architecture diagram illustrating the Generator-Retriever-Generator (GRG) approach, which combines document retrieval techniques and large language models to generate contextual documents and retrieve relevant information for answering questions.}
  \label{fig:figure_label}
\end{figure}
\subsection{Retriever Only}
The Retrieval-Only~\cite{lee2020learning} approach seeks to reformulate open-domain question answering as a phrase retrieval problem, eliminating the need for processing documents during inference. Previous work has explored retrieval models that heavily rely on sparse representations, such as TF-IDF or BM25~\citep{karpukhin2020dense} to retrieve relevant phrases or sentences. However, these models often underperform compared to retriever-reader approaches. Recent work has then focused on learning dense representations of phrases alone, leading to stronger performance in open-domain question answering. This involves training models using reading comprehension tasks and employing negative sampling techniques. \citet{seo2019real} proposed a phrase retrieval approach in which they independently encode the representations of phrases and questions. They then utilize a similarity search over the encoded phrase representations to identify the correct answer.

Table~\ref{tab:advantages_disadvantages} presents the advantages and disadvantages of each of the 4 approaches in question answering systems. Retrieve-Reader leverages external knowledge and document-based context, but there is a possibility of missing relevant documents and dependency on retrieval performance. Generate-Reader offers flexibility and adaptability in generating answers, but it requires substantial computational power, and the generated answers may not always be accurate. Retrieve-Generate balances retrieval and generation, enhancing recall but increasing computational complexity. Retrieve-Only leverages external knowledge and document-based context, but it has limitations in handling complex queries and lacks flexibility. Generator-Retriever-Generator provides contextual relevance, improved accuracy, and adaptability, but it comes with increased computational complexity and the challenge of balancing quality and diversity. These considerations play a crucial role in designing effective question-answering systems.


\section{Method}

Figure~\ref{fig:figure_label} presents an architectural diagram depicting the GRG approach and its sequential process.
It comprises three integral components: (i) a \textit{large language model (LLM) for document generation}, (ii) a \textit{dual-encoder network for document retrieval}, and (iii) a \textit{second 
large language model for answer generation}.  
In the following sections, we discuss each component in detail and outline our training methodology.

\subsection{Document Generation}

Few-shot information extraction tasks aim to recognize novel relations and extract relevant information from unstructured text with limited annotated instances~\citep{han-etal-2021-exploring}. Traditional information extraction methods struggle with data scarcity and often face challenges in identifying emerging relation types and their associated entity pairs. To overcome this issue, few-shot learning techniques leverage a small number of labeled samples to generalize to unseen instances~\cite{liu2018learning}.

For our case, generating informative and contextually rich background documents can be used as a few-shot technique when the power of language models, particularly, InstructGPT~\cite{ouyang2022training}, is harnessed.
GRG then uses InstructGPT to generate context by providing an input prompt. For few-shot information extraction, a suitable prompt structure could be: "Generate a background document to answer the given question: [question placeholder]". By substituting the "question placeholder" with the actual question, we instruct the model to generate a document that contains pertinent information for answering the question. Utilizing InstructGPT, we generate informative and contextually rich documents that provide relevant information for answering a given question. These generated documents are then included in the collection of evidence documents $\mathcal{D}$.
\subsubsection{Vector Index Retrieval}
We propose a vector-based retrieval \cite{Liu_LlamaIndex_2022} method to increase relevance of knowledge in generated documents using the \textit{Vector Index Retriever} \cite{ xiao2022distill}. This approach leverages vector representations and the \textit{ Vector Store Index}\footnote{\url{https://github.com/jerryjliu/llama_index}} to efficiently retrieve documents based on their similarity to the input question. The \textit{Vector Index Retriever} is crucial to our information retrieval pipeline. It utilizes the \textit{ Vector Store Index}, which stores vector representations of documents generated by a large language model. We capture each document's semantic and contextual information by encoding each document with a high-dimensional vector. In the retrieval process, the \textit{Vector Index Retriever} employs a similarity-based approach to identify the most relevant documents. Given a question, it retrieves a pre-specified number of \textit{top k} results with the highest similarity scores. The \textit{k} parameter can be adjusted to balance the precision and efficiency. We describe the details of each step below.

\textbf{Step 1: Generate Documents.} We first generate 10 to 50 contextual documents $D_{G}$ for each question $q \in \mathcal{Q}$ using InstructGPT. Here, $\mathcal{Q}$ represents the set of questions in the dataset.

\textbf{Step 2: Encode each Document.} Using GTR-T5-large/MiniLM-L6 \cite{reimers2019sentencebert,ni2021large} language model, we encode each document $d_i$, resulting in a 768/384-dimensional vector $\textbf{e}_i$ per document.

\textbf{Step 3: Vector Index Representation.} We store all the embedding vectors $\{\textbf{e}_i\}_{i=1}^{|Q|}$ using the \textit{Vector Store Index}. This allows for efficient retrieval of documents based on their similarity to the question.

\textbf{Step 4: Selection of Generated Documents.} After storing the encoded documents, we utilize the \textit{Vector Index Retriever} to process the question and select up to \textit{top k} (2 or 5 in our experiments) the most relevant documents with a high cosine similarity score threshold
The cosine similarity score is calculated between the encoded question vector and the vectors of the stored documents:
\[
\text{{Cosine Similarity Score}}(\mathbf{q}, \mathbf{d}_i) = \frac{{\mathbf{q} \cdot \mathbf{d}_i}}{{\|\mathbf{q}\| \cdot \|\mathbf{d}_i\|}}
\]

where $\mathbf{q}$ represents the encoded question vector and $\mathbf{d}_i$ represents the vector of the $i$-th stored document. 

By comparing the cosine similarity scores of the question vector with the vectors of the stored documents, we can identify the most relevant documents that have high similarity to the question. In this case, we retrieve the top 5 documents with similarity above the specified threshold of 0.7.
By following these steps, our approach enables effective retrieval of generated contextual documents for open-domain question-answering, specifically selecting documents with high similarity to the question and, thus ones that are likely to contain the correct answer. This retrieval process leverages vector representations and similarity-based techniques to prioritize the most relevant and informative documents.

\subsection{Document Retriever}
The retriever module plays a crucial role in our question-answering model. Given a collection of evidence documents $\mathcal{D_{R}} = \{\boldsymbol{d}_1, \ldots, \boldsymbol{d}_M\}$ and a question $\boldsymbol{q}$, its goal is to select a subset of the documents $\mathcal{Z} \subset \mathcal{D_{R}}$ that are most relevant to the question. This subset of documents will be used for further processing and answer generation. For this, our retriever model is based on EMDR (End-to-end training of Multi-Document Reader and Retriever) \citep{singh2021end}, which is a dual-encoder network \citep{vaswani2017attention} consisting of two separate encoders: $f_q$ for encoding the question and $f_d$ for encoding the evidence documents. Each encoder takes a sequence (question or document) as input and produces its fixed-size vector representation. To quantify the relevance or similarity between a question $q$ and an evidence document $d_i$, we compute their respective encoded vectors using the encoders $f_q$ and $f_d$. The retrieval score is then determined by taking the dot product between these vectors: 

\begin{align}
\text{score}(q, d_i; \Phi) = \text{enc}(q; \Phi_q) \cdot \text{enc}(d_i; \Phi_d)
\end{align}

Where $\text{enc}(q; \Phi_q)$ and $\text{enc}(d_i; \Phi_d)$ represent the encoded vectors of the question and document, respectively, with $\Phi$ denoting the retriever parameters.
By calculating the dot product, we capture the similarity between the question and document, with higher scores indicating stronger relevance.
Based on the retrieval scores, we select the top-$k$ documents from the collection $\mathcal{D_{R}}$ for a given question $q$ which are indicated as $\mathcal{Z} = {z_1, \ldots, z_k}$. 

\subsection{Generation Model}

Our generator is based on a model from the LLaMA family - a collection of open-source language models pretrained on trillions of tokens using publicly available datasets, which achieve state-of-the-art performance on many benchmarks. The generator model takes as input a question $q$ and a set of retrieved and generated documents to generate an answer.

Each retrieved document $z_i$ and generated document $d_i$ are concatenated with the question. We use the newline character (\textbackslash n) as a delimiter to ensure  separation between the documents. Additionally, we include the \texttt{</s>} token at the end of each utterance as an end-of-turn token, which indicates the completion of each input segment.

The input to our generator model is then represented as follows:
\[
\text{{input}} = [q, z_{im}, \text{{\textbackslash n}}, d_{im}, \text{{\textbackslash n}}, \texttt{</s>}]
\]

The LLaMA language model uses a novel loss function called cosine loss that helps the model to better distinguish between similar words and improve its accuracy. The cosine loss is defined as follows:
$$\mathcal{L}_{cos} = -\frac{1}{N}\sum_{i=1}^{N} \log \frac{\exp(\cos(\mathbf{h}_i,\mathbf{t}_i)/\tau)}{\sum_{j=1}^{N}\exp(\cos(\mathbf{h}_i,\mathbf{t}_j)/\tau)}$$

where $\mathbf{h}_i$ is the hidden state of the $i$-th token in the sequence and $\mathbf{t}_i$ is the target embedding for that token. $\tau$ is a temperature parameter that controls the sharpness of the distribution.

By incorporating the question, retrieved documents, and generated documents, our generator model can generate contextually informed answers tailored to the specific question and the available input information.

\section{Experimental Settings}

\subsection{Datasets}

The evaluation is conducted on several datasets, following the same experimental setup as in \cite{yu2022generate,izacard2020leveraging,lee2019latent}. 
We consider the following datasets:

\begin{itemize}
    \item \textbf{\texttt{NaturalQuestions}} \cite{kwiatkowski2019natural}: This dataset consists of questions corresponding to Google search queries. Natural Questions (NQ)\footnote{NQ (Retriever): \url{https://ai.google.com/research/NaturalQuestions/download} and NQ (Generator): \url{https://drive.google.com/drive/folders/1DNjTTOLKi24wohJKu1Z-v6b4izfymlLu}} was generated from real Google search queries, and the answers are spans within Wikipedia articles. The NQ dataset consists of around 79,168 examples in the training set, 8,757 examples in the development set, and 3,610 examples in the test set.
   
    \item  \textbf{\texttt{TriviaQA}} \cite{joshi2017triviaqa}: This dataset contains questions collected from trivia and quiz-league websites. For open-domain question answering, we use the unfiltered version of the dataset. TriviaQA\footnote{TQA (Retriever): \url{http://nlp.cs.washington.edu/triviaqa/ } and TQA (Generator): \url{https://drive.google.com/drive/folders/1DNjTTOLKi24wohJKu1Z-v6b4izfymlLu}} is a collection of trivia questions sourced from trivia and quiz-league websites. The dataset includes 78,785 examples in the training set, 8,837 examples in the development set, and 11,313 examples in the test set.

    \item  \textbf{\texttt{WebQ}} \cite{berant2013semantic}: WebQuestions (WebQ) \footnote{ WebQ (Retriever): \url{https://github.com/google-research/language/tree/master/language/orqa} and  WebQ (Generator): \url{https://drive.google.com/drive/folders/1DNjTTOLKi24wohJKu1Z-v6b4izfymlLu}} consists of questions obtained using the Google Suggest API, with the answers being entities from Freebase. The dataset contains approximately 3,417 examples in the training set, 361 examples in the development set, and 2,032 examples in the test set. 
\end{itemize}

To evaluate the performance of our model, we employ the exact match (EM) score, following 
\citet{zhu2021retrieving,yang2019end,chen2020open}. The EM score measures the correctness of an answer by comparing its normalized form to the acceptable answer list. Through these evaluations, we aim to assess the effectiveness of the GRG model in the domain of open-domain question answering.

We adopt the train/dev/test splits that have been previously used in the open-domain QA setting, as employed by~\citet{izacard2020leveraging} and~\citet{karpukhin2020dense}.  Table~\ref{table:dataset} presents the statistics of the dataset sizes, including the training, development, and test sets.
We note that all our models are trained exclusively on the training data, and we did not include the development data in our training process. Therefore, the performance numbers reported in the paper for the dev and test data are independent of the training data. We split the training data, allocating 90\% for model training and the remaining 10\% for testing purposes.
\begin{table}[t]
\small
\caption{Datasets' statistics.}
\centering
\begin{tabular}{l c c c c}
  \toprule
  \textbf{Dataset} & \textbf{Train}  & \textbf{Dev} & \textbf{Test} \\
  \midrule
  WebQ & \phantom{0}3,417  & \phantom{00}361 & \phantom{0}2,032 \\
  NQ & 79,168 & 8,757 & \phantom{0}3,610 \\
  TQA & 78,785  & 8,837 & 11,313 \\
  \bottomrule
\end{tabular}

\label{table:dataset}
\end{table}

\begin{table}[t]
\centering
\caption{Training and Hyperparameter Settings for LLaMa-7B}

\small
\begin{tabular}{llll}
\toprule
\textbf{Parameter} & \textbf{Value} & \textbf{Parameter} & \textbf{Value} \\
\midrule
Attention heads & 32 & Optimizer & AdamW \\
n layers & 32 & beta1 & 0.9 \\
dimension & 4096 & beta2 & 0.999 \\
Hardware & A100 and A40 & epsilon & 1e-08 \\
Batch Size & 4 & gradient accumulation steps & 8 \\
CPU & 100 & learning rate & 2e-05 \\
weight decay & 0.0 & max grad norm & 1.0 \\
train batch size & 4 & eval batch size & 4 \\
warmup ratio & 0.03 & Warm-up Steps & 2,000 \\

\bottomrule
\end{tabular}
\label{tab:trainingsetting}
\vspace{-.5cm}
\end{table}
\subsection{Choice of Document Number} In our approach, we used only 2 or 5 documents during the generator process due to computational limitations and the extensive training time required for the LLaMA model. As~\citet{izacard2020leveraging} reported, training the T5 model using 100 documents necessitates considerable computational resources, such as 64 Tesla V100 32GB GPUs running for approximately one day. While increasing the number of documents can enhance model performance~\citep{izacard2020leveraging}, it incurs significant costs regarding memory consumption and training time, which should be carefully considered, especially, in the current trend towards GreenAI~\citep{yu2022generate}. 
\subsection{Experimental Setup}
In this section, we describe the experimental setup for training the LLaMA model using the DeepSpeed framework \citep{deepspeed_zero}. DeepSpeed provides techniques and automated parameter tuning to optimize training efficiency and memory utilization. We customized the training process using DeepSpeed's configuration options. Firstly, we enabled mixed precision training with bfloat16 (bf16) precision to accelerate training while maintaining accuracy. The AdamW optimizer was selected, and its hyperparameters were determined automatically by DeepSpeed. To control the learning rate, we employed the WarmupDecayLR scheduler. The LLaMA model is based on the transformer architecture \citep{vaswani2017attention} widely used in large language models. We utilize the LLaMa-7B model as our backbone for implementing GRG. The training and hyperparameter settings for LLaMa-7B are summarized in Table~\ref{tab:trainingsetting}.

For memory consumption and speed optimization, we utilized DeepSpeed's zero optimization stage 3, offloading the optimizer state and model parameters to the CPU with pinned memory. Additional hyperparameters were set, including gradient accumulation steps (8 steps), gradient clipping (determined automatically), and batch size (value of 4). This experimental setup aimed to achieve efficient training and optimal performance of our LLaMA model.

In addition to the DeepSpeed experimental setup described above, we conducted an additional experiment using the LoRA technique \citep{hu2021lora} for fine-tuning our LLaMA model. LoRA, which stands for "Low-Overhead Representation Adaptation," is a method that allows for the efficient fine-tuning of large language models. For this experiment, we followed a slightly different approach. Instead of recreating the entire model from scratch, we generated a fine-tuning file that would be applied to the base Llama model. This approach significantly reduces computational overhead and makes the fine-tuning process more efficient, even on modest hardware.

Our proposed model and relevant baselines are implemented using PyTorch~\citep{paszke2019pytorch} on a cluster of machines equipped with 100 CPUs, 400GB of physical memory, and a combination of 4 A40 and 4 A100 GPUs for our experiments.

\begin{table}[t]
\centering
\caption{Performance Comparison of GRG  Approach and Baseline Models on TriviaQA, WebQ, and NQ Datasets.}
\begin{adjustbox}{width=0.5\textwidth,center}
\begin{tabular}{l|cc|cccccc}
\toprule
\multirow{2}{*}{Models} & \# reader & {\# docu-} & \multicolumn{2}{c}{TriviaQA} & \multicolumn{2}{c}{WebQ} & \multicolumn{2}{c}{NQ}  \\
& parameters & ments& dev& test & dev& test & dev& test  \\  
\midrule
\multicolumn{8}{l}{\textit{*baselines with retrieving from Wikipedia; all numbers reported by existing papers}} \\
\textbf{BM25 + BERT}~\cite{lee2019latent}& 220M & 5 & 47.2  & 47.1 & 27.1 & 21.3 &  24.8 & 26.5 \\
\textbf{REALM}~\citep{guu2020retrieval} & 330M & 5 & -& - &  -& 40.7 &  38.2 & 40.4   \\
\textbf{DPR}~\citep{karpukhin2020dense} & 110M & 100 & -& 56.8 &  -& 41.1 &  -& 41.5  \\
\textbf{RAG}~\citep{lewis2020retrieval} & 400M & 10 &-& 56.1 &-& 45.2 &-& 44.5  \\

\textbf{FiD-l}~\cite{yu2022generate}& 770M & 10 &-& 61.9 &-& 48.1 &-& 46.7  \\
\textbf{FiD-xl}~\cite{yu2022generate}  & 3B & 10 &-& 66.3 &-& 50.8 &-& 50.1 \\
\textbf{FiD-xl}~\cite{yu2022generate}  & 3B & 10 &-& 70.1 &-& 53.6 &-& 45.0  \\
\textbf{FiD}~\citep{izacard2020leveraging} & 770M & 100 &-& 67.6 &-& 50.5 &-& 51.4   \\
\textbf{EMDR}~\citep{singh2021end}& 440M & 50 &71.1& 71.4 &49.9& 48.7 &50.4& 52.5  \\
\textbf{RFiD-large}~\citep{wang2023rfid}& 990M & 100 &72.7  & 72.6 & -& - & 52.5 &  54.3  \\
\midrule
\multicolumn{8}{l}{\textit{*baselines with phrase retrieval; all numbers reported by existing papers}} \\
\textbf{DensePhrases}~\citep{lee2020learning}& 110M & 50 &- &34.4 & -& 17.3 & - & 14.5  \\
\textbf{DensePhrases}~\citep{lee2021learning}& 110M & 50 &- &53.5 & -& 41.5 & - & 41.3  \\
\midrule
\multicolumn{8}{l}{\textit{*baselines with generated documents; all numbers reported by existing papers}} \\
\textbf{GenRead (FiD-l)}~\cite{yu2022generate}  & 770M & 10 &-& 67.8 &-& 51.5 &-& 40.3  \\
\textbf{GenRead (FiD-l)}~\cite{yu2022generate}  & 770M & 10 &-& 70.2 &-& 53.3 &-& 43.5  \\
\textbf{GenRead (FiD-xl)}~\cite{yu2022generate}  & 3B & 10 &-& 69.6 &-& 52.6 &-& 42.6 \\
\textbf{GenRead (FiD-xl)}~\cite{yu2022generate}  & 3B & 10 & -& 71.6 &-& 54.4 &-& 45.6  \\
\midrule
\multicolumn{8}{l}{\textit{*baselines with generated and retrieved documents}} \\
\textbf{COMBO}~\cite{zhang2023merging} &  3B & 2 & - & 74.6  & - &\textbf{54.2}  &- & 53.0  \\

\midrule
\multicolumn{8}{l}{\textit{*our proposed method by combining generated and retrieved documents}} \\
\textbf{GRG (LoRA)} & 1.2B & 2  & 67.6 & 69.1 & 48.6& 45.2 &  50.8 &  49.1 \\
\textbf{GRG (LoRA)} & 1.2B & 5  & 69.4 & 70.8 & 50.6 &   42.9 & 54.8 & 53.4    \\

\textbf{GRG} & 7B & 2  & \textbf{76.4} & \textbf{75.7} & \textbf{52.0}  & 53.6  & \textbf{55.4} & \textbf{57.4}  \\
\textbf{GRG} & 7B & 5  &  \textbf{77.1} & \textbf{76.8} & \textbf{55.8}&  \textbf{56.0} & \textbf{56.2} & \textbf{58.5 } \\
\bottomrule
\end{tabular}
\end{adjustbox}
\label{tab:result}
\end{table}

\begin{table}[h]
\centering
\caption{Recall@K scores for document retrieval using our approach equipped with GTR-T5-large and MiniLM-L6 models on TQA, NQ, and WebQ datasets.}

\begin{tabular}{c|cc|cc|cc}
\toprule
\multirow{2}{*}{Models} & \multicolumn{2}{c}{TQA} & \multicolumn{2}{c}{NQ}  & \multicolumn{2}{c}{WebQ} \\
 & dev & test & dev & test & dev & test \\
\midrule
\textbf{MiniLM-L6} &  76.1 & 76.7 & 58.6 & 60.3 &67.0& 60.1\\
\textbf{GTR-T5}  &  \textbf{78.5} &  \textbf{79.2} &  \textbf{62.2}&  \textbf{63.9}& \textbf{72.6} & \textbf{68.1}\\
\bottomrule
\end{tabular}
\label{tab:Recall_K}
\end{table}
\section{Results}

We present in this section the experimental results, which are divided into three subsections: Results of Open-Domain QA, Results of document generation, and the Ablation study. The document generation analysis aims to evaluate the effectiveness of our document retrieval method in generating relevant and informative documents for answering open-domain questions. In the ablation study, we investigate the impact of different factors (top-k answers, architecture components, and zero-shot strategy) on the performance.

\subsection{Results of Open-Domain QA}

This section presents the results of the proposed \textbf{GRG} approach, which combines generated and retrieved documents for question answering. The results of the experiments are shown in Table~\ref{tab:result} using EM score.
We compare the performance of \textbf{GRG} against several baselines and existing state-of-the-art models on three benchmark datasets: \texttt{TriviaQA}, \texttt{WebQ}, and \texttt{NQ}. 
We first compare \textbf{GRG} against baseline models that utilize document retrieval from Wikipedia. These baselines include \textbf{BM25 + BERT}~\cite{lee2019latent}, \textbf{REALM}~\citep{guu2020retrieval}, \textbf{DPR}~\citep{karpukhin2020dense}, \textbf{RAG}~\citep{lewis2020retrieval}, \textbf{FiD-l}~\cite{yu2022generate}, \textbf{FiD-xl}~\cite{yu2022generate}, \textbf{FiD}~\citep{izacard2020leveraging}, \textbf{EMDR}~\citep{singh2021end}, \textbf{DensePhrases} models \citep{lee2020learning,lee2021learning}, and \textbf{RFiD-large}~\citep{wang2023rfid}. The numbers reported for these baselines are taken directly from their respective papers. 
\textbf{GRG} consistently outperforms most of the baseline models across all datasets. Specifically, \textbf{GRG} achieves significant improvements over \textbf{BM25 + BERT} (29.9\% improvement on \texttt{TriviaQA} dev set) and (29.7\% improvement on \texttt{TriviaQA} test set), \textbf{REALM} (15.3\% improvement on \texttt{WebQ} test set), \textbf{DPR} (14.9\% improvement on \texttt{WebQ} test set), \textbf{FiD} (7.1\% improvement on \texttt{NQ} test set), and \textbf{RAG} (14.0\% improvement on \texttt{NQ} test set), demonstrating the effectiveness of the combined generated and retrieved documents' based approach.
Next, we compare \textbf{GRG} against \textbf{DensePhrases} models \citep{lee2020learning,lee2021learning} that employ phrase retrieval. \textbf{DensePhrases} has been shown to perform well in question-answering tasks. However, \textbf{GRG} approach surpasses the performance of \textbf{DensePhrases} across all datasets. On \texttt{TriviaQA} dev set, \textbf{GRG} achieves a 23.3\% improvement over \textbf{DensePhrases}~\citep{lee2020learning}, and on \texttt{WebQ} test set, it has an 14.5\% improvement over \textbf{DensePhrases} \citep{lee2021learning}.

Next, we evaluate the performance of \textbf{GRG} against \textbf{GenRead}~\cite{yu2022generate} models that only generate documents. \textbf{GenRead} models have shown promising results in generating informative documents. Still, our approach consistently outperforms \textbf{GenRead} regarding question answering accuracy on all the datasets. On \texttt{TriviaQA} dev set, \textbf{GRG} achieves a 7.3\% improvement over \textbf{GenRead (FiD-l)}, and on \texttt{WebQ} test set, it has a 2.1\% improvement over \textbf{GenRead (FiD-l)}. 

Finally, we discuss the performance of \textbf{GRG} with varying configurations. We evaluate \textbf{GRG} with two numbers of generated documents (2 and 5) using LoRA. Additionally, we report the performance of \textbf{GRG} without LoRA, utilizing the same number of generated documents. 
On the \texttt{TriviaQA} dev set, \textbf{GRG} achieved 76.4\% accuracy when using 2 generated documents, which rose to 77.1\% for the case of 5 generated documents. The performance on the \texttt{WebQ} test set of the model is 52.0\% accuracy with 2 generated documents, increasing to 55.8\% with 5 generated documents. Lastly, on the \texttt{NQ} test set, the model achieved an accuracy of 55.4\% with 2 generated documents and showed a slight improvement to 56.2\% when 5 generated documents were utilized.
\textbf{GRG} outperforms all of the baselines on all three datasets. When applied on \texttt{TriviaQA}, \textbf{GRG} achieves an exact match score of 76.8, which is a +5.2 improvement over the previous state-of-the-art (\textbf{GenRead}). Testing on On \texttt{WebQ}, we see that our model reaches an exact match score of 56.0, which is a +1.6 improvement over the previous state-of-the-art (\textbf{RFiD-large}). On the last dataset, \texttt{NQ}, \textbf{GRG} achieves an exact match score of 58.5, which is a +4.2 improvement over the previous state-of-the-art (\textbf{GenRead}).  we also compare our \textbf{GRG} approach with the \textbf{COMBO} model, which also utilizes a blend of generated and retrieved documents, and has shown notable performance, particularly on the \texttt{TriviaQA} and \texttt{WebQ} datasets. Notably, it achieves an exact match score of 74.6 on the \texttt{TriviaQA} test set and 54.2 on the \texttt{WebQ} test set, representing state-of-the-art performance on these datasets. However, when compared to \textbf{GRG}, our approach still shows superior results. Specifically, \textbf{GRG} achieves a higher exact match score of 76.8 on \texttt{TriviaQA} and 56.0 on \texttt{WebQ}, surpassing COMBO by 2.2 and 1.8 points, respectively. This suggests that while COMBO's strategy is effective, the methodologies employed in \textbf{GRG} allow for even more precise question-answering capabilities. 

Our results demonstrate that \textbf{GRG} performs better than all the baselines and state-of-the-art models across all the datasets. Including generated and retrieved documents enables \textbf{GRG} to capture a wider range of relevant information, improving QA accuracy. Notably, \textbf{GRG} with 5 generated documents consistently outperforms \textbf{GRG} with 2 generated documents, suggesting the benefit of incorporating more diverse generated content. 

\subsection{Evaluating Document Generation}
In this section, we present the experimental results of our document retrieval approach for document generation using the \textbf{GTR-T5-large} and \textbf{MiniLM-L6} models. We computed the Recall@K of retrieving the documents containing the true answer for each question same as in~\cite{sachan2022improving}. To ensure a fair comparison and consistent evaluation, we utilized the same dataset as in \cite{yu2022generate}. The choice of using the same dataset was motivated by the fact that the generated context from the \textbf{InstructGPT} model may significantly differ for every request. We measured the Recall@K of our document retrieval method by calculating the percentage of questions for which the retrieved document contained the true answer. 
These accuracy results highlight the effectiveness of our vector index retrieval approach in identifying relevant documents for answering open-domain questions. \textbf{GTR-T5-large} model, with its higher-dimensional vector encoding, exhibits better performance compared to the \textbf{MiniLM-L6} model and the approach proposed by \citet{yu2022generate}.
Table~\ref{tab:Recall_K} presents Recall@K scores for three question answering datasets: \texttt{TQA}, \texttt{NQ}, and \texttt{WebQ}. The \textbf{MiniLM-L6} model achieves scores ranging from 58.6\% to 76.7\% across the datasets, while the \textbf{GTR-T5-large} model outperforms it with scores ranging from 62.2\% to 79.2\% for the respective datasets.

\section{Ablation studies}

\subsection{Zero-Shot Open-Domain QA} Table~\ref{tab:zero_results} shows the results of a zero-shot open-domain question answering (QA) evaluation, where different models are assessed without any external documents. These models, including \textbf{FLAN}, \textbf{GLaM}, \textbf{Chinchilla}, \textbf{Gopher}, \textbf{InstructGPT}, \textbf{GPT-3}, and \textbf{LLaMA} \cite{rae2021scaling,wei2021finetuned,du2022glam,roberts2020much,ouyang2022training,touvron2023llama}, possess varying parameter sizes and have been trained on large-scale corpora, enabling them to capture extensive world knowledge. When examining the performance of each model on answering questions from the \texttt{TQA}, \texttt{NQ}, and \texttt{WebQ} datasets, we observe notable variations. \textbf{LLaMA}, with its 7B parameters, stands out by achieving remarkable results in zero-shot QA. Despite the relatively smaller parameter size, \textbf{LLaMA} demonstrates the ability to effectively leverage the knowledge embedded within its parameters, showcasing its potential as a powerful tool for zero-shot question answering tasks. Models like \textbf{InstructGPT} and \textbf{GPT-3}, with larger parameter sizes (175B), also demonstrate competitive performance. \textbf{InstructGPT} achieves a high accuracy of 57.4\% on the \texttt{TQA} dataset and performs consistently well across the other datasets. \textbf{GPT-3} also achieves competitive results.

\subsection{Impact of Architecture Components}

\begin{figure}[htbp]
  \centering
  \includegraphics[width=0.3\textwidth]{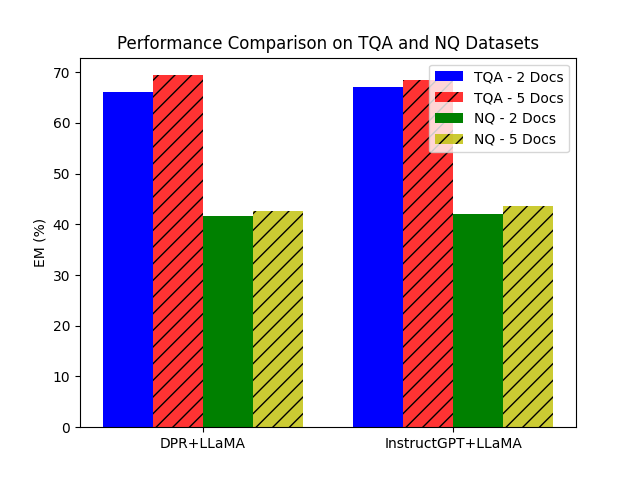}
  \caption{Performance Comparison (EM) of DPR+LLaMA and InstructGPT+LLaMA models on TQA and NQ. 
  }
  \label{fig:tqa_nq_performance}
\end{figure}

We now evaluate the performance of each component used in our approach, specifically the retriever and the generator, when combined with \textbf{LLaMA}. The goal is to understand the individual contributions of these components on the overall performance. We compare the results on the \texttt{TQA} and \texttt{NQ} datasets using different combinations of models. Figure \ref{fig:tqa_nq_performance} shows the performance comparison of \textbf{DPR+LLaMA} and \textbf{InstructGPT+LLaMA} models on \texttt{TQA} and \texttt{NQ} datasets.

On the \texttt{TQA} dataset, the \textbf{InstructGPT+LLaMA} model demonstrated an EM score of 67.1\% and 70.1\% on the development and test sets, respectively, when trained with 2 documents. Upon using 5 documents for training, the performance improved to 68.4\% and 71.8\% on the development and test sets, respectively. Shifting the focus to the \texttt{NQ} dataset, the \textbf{InstructGPT+LLaMA} model showed competitive performance, achieving an EM score of 42.1\% on the development set and 42.0\% on the test set with 2 documents. Increasing the number of training documents to 5 resulted in a modest improvement, with EM scores of 43.6\% on the development set and 44.5\% on the test set. These findings indicate that incorporating more documents during training can positively impact model performance. There may be however a diminishing return in terms of accuracy improvement. As a result, striking a careful balance between the number of training documents and the resulting performance may be crucial to optimize computational resources and training time.

\begin{table}[t]
\centering
\caption{ Comparative Performance of Language Models in Zero-Shot Open-Domain QA.}
\small
\setlength{\tabcolsep}{1mm}{
\begin{tabular}{c|c|ccc}
\toprule
Models & parameters & TQA & NQ & WebQ \\
\midrule
\textbf{FLAN}  & 137B & 56.7 & 20.7 & - \\
\textbf{GLaM}  & 64B & - & 21.5 & 19.0\\
\textbf{Chinchilla}  & 70B & 55.4 & 16.6 & -\\
\textbf{PaLM}  & 540B & - & \textbf{21.2}& 10.9\\
\textbf{Gopher} & 280B & 43.5   & 10.1 & 35.6\\
\textbf{InstructGPT} & 175B & \textbf{57.4}  & 19.5& 19.9\\
\textbf{GPT-3} & 175B  & 49.2  & 14.6 & 14.4\\
\textbf{LLaMA} & 7B & 50.0   & 16.8& \textbf{28.8}\\
\bottomrule
\end{tabular}}

\label{tab:zero_results}
\end{table}

\subsection{Comparative Analysis  GRG (LoRA) Models}
We present additional experimental results for document generation using the GRG (LoRA) and GRG models. The performance of these models is evaluated on the TQA and NQ datasets, and the results are summarized in Table~\ref{tab:documentgeneration}.

\begin{table}[t]
\centering
\caption{F1 scores for document generation using GRG and GRG (LoRA) models.}
\small
\setlength{\tabcolsep}{1.4mm}{
\begin{tabular}{c|c|cc|cc}
\toprule
\multirow{2}{*}{Models} &  {\# docu-} & \multicolumn{2}{c}{TQA} & \multicolumn{2}{c}{NQ}  \\

 & ments & dev & test & dev & test  \\
\midrule

 GRG (LoRA)& 2& 75.6  & 78.8 &  60.4 &  59.5 \\
GRG (LoRA)&5 & 79.7 & 80.4 & 63.9 &  61.7 \\
GRG  & 2& 84.0 & 83.8 & 64.6 & 65.0\\
GRG  &5&  84.6 & 84.7 & 65.4 & 66.1  \\

\bottomrule
\end{tabular}}
\label{tab:documentgeneration}
\end{table}
Table~\ref{tab:documentgeneration} displays the F1 scores obtained by the GRG (LoRA) and GRG models when generating documents for the TQA and NQ datasets. The models are evaluated on both the development and test sets.

For GRG (LoRA) model, the results indicate that increasing the number of documents from 2 to 5 leads to improved performance on both datasets. On the TQA dataset, the F1 score increases from 75.6 to 79.7 on the development set and from 78.8 to 80.4 on the test set when moving from 2 to 5 documents. Similarly, on the NQ dataset, the F1 score improves from 60.4 to 63.9 on the development set and from 0.595 to 0.6173 on the test set.

The GRG model also demonstrates competitive performance in document generation. With 2 documents, the model achieves an F1 score of 84.05 on the TQA development set and 83.8 on the test set. On the NQ dataset, the F1 score is 64.6 on the development set and 65.0 on the test set. Increasing the number of documents to 5 further enhances the performance, with F1 scores of 84.6 and 84.7 on the TQA development and test sets, respectively, and 65.4 and 66.1 on the NQ development and test sets, respectively.

\begin{table}[t]
\centering
\caption{Performance comparison of DPR and InstructGPT models on the TQA dataset.}
\small
\setlength{\tabcolsep}{1.4mm}{
\begin{tabular}{c|cc|cc}
\toprule
\multirow{2}{*}{Models} & \multicolumn{2}{c}{Development Set} & \multicolumn{2}{c}{Test Set} \\
& 2 Docs & 5 Docs & 2 Docs & 5 Docs \\
\midrule
DPR+LLaMA & 66.0\% & 69.4\% & 66.8\% & 69.3\%\\
InstructGPT+LLaMA & 67.1\% & 68.4\% & 70.1\% & 71.8\% \\
\bottomrule
\end{tabular}}
\label{tab:tqa_performance}
\end{table}

\begin{table}[t]
\centering
\caption{Performance comparison of DPR and InstructGPT models on the NQ dataset.}
\small
\setlength{\tabcolsep}{1.4mm}{
\begin{tabular}{c|cc|cc}
\toprule
\multirow{2}{*}{Models} & \multicolumn{2}{c}{Development Set} & \multicolumn{2}{c}{Test Set} \\
& 2 Docs & 5 Docs & 2 Docs & 5 Docs \\
\midrule
DPR+LLaMA & 41.7\% & 42.6\% & 41.2\% & 42.6\%\\
InstructGPT+LLaMA & 42.1\% & 43.6\% & 42.0\% & 44.5\% \\
\bottomrule
\end{tabular}}

\label{tab:nq_performance}
\end{table}

\begin{table}[h!]
\centering
\caption{Performance Comparison (EM and F1) Scores of GRG for different top-k values on NQ and TQA datasets}
\begin{adjustbox}{width=0.5\textwidth,center}
\begin{tabular}{l|ll|ll|ll|ll}
\hline
\multirow{2}{*}{\textbf{Top-k}} & \multicolumn{2}{c|}{\textbf{NQ}} & \multicolumn{2}{c|}{\textbf{TQA}} & \multicolumn{2}{c|}{\textbf{NQ}} & \multicolumn{2}{c}{\textbf{TQA}} \\
\cline{2-9}
& \textbf{Dev EM} & \textbf{Test EM} & \textbf{Dev EM} & \textbf{Test EM} & \textbf{Dev F1} & \textbf{Test F1} & \textbf{Dev F1} & \textbf{Test F1} \\
\hline
1 & 56.2 & 58.5 & 77.1 & 76.8 & 65.4 & 66.1 & 84.6 & 84.7 \\
2 & 66.1 & 67.3 & 79.9 & 80.0 & 72.9 & 73.3 & 86.6 & 86.7 \\
3 & 68.8 & 70.3 & 81.3 & 81.4 & 75.3 & 75.9 & 87.6 & 87.8 \\
4 & 70.6 & 71.9 & 82.1 & 82.1 & 76.8 & 77.2 & 88.3 & 88.3 \\
5 & \textbf{71.6} & \textbf{72.8} & \textbf{82.6} & \textbf{82.6} & \textbf{77.7} & \textbf{78.4} & \textbf{88.7} & \textbf{88.8} \\
\hline
\end{tabular}
\end{adjustbox}

\label{tab:combined_scores}
\end{table}

\subsection{Impact of top-k Answer on Performance}
We finally analyze the impact of different top-k values on the performance of our proposed approach. Table~\ref{tab:combined_scores} presents the EM and F1 scores for different top-k values on \texttt{NQ} and \texttt{TQA} datasets. We observe that as the top-k value increases, the EM scores consistently improve. For example, on the \texttt{NQ} dataset, the EM score increases from 56.3\% at top-1 to 71.6\% at top-5. Similarly, on \texttt{TQA}, the EM score increases from 76.2\% at top-1 to 82.6\% at top-5.

\section{Limitations}
 This study acknowledges the following potential limitations:
\begin{enumerate}
    \item Generated Document Quality: The performance of our approach depends on the accuracy and relevance of the documents generated by the language model. Despite extensive training, there can be instances of inaccurate or irrelevant information.
    \item Large language models can be computationally intensive and time-consuming, especially for complex queries. This can pose scalability challenges when processing a large number of queries or with limited computing resources. More details are in Appendix \ref{cost_anaysis}.
\end{enumerate}

\section{Conclusions}
In this paper, we proposed a Generator-Retriever-Generator approach for improving open-domain question answering systems. By combining generated and retrieved documents, we achieved significant performance gains across multiple benchmark datasets. Our experiments demonstrate that GRG outperforms existing baselines in terms of accuracy and efficiency. The results indicate also the effectiveness of incorporating both generated and retrieved documents in the reading process, leveraging the combined strengths of language models and retrieval systems. 

Future work should focus on improving the accuracy of the document retrieval approach, potentially through the use of more advanced retrieval models or by incorporating additional contextual information. Further, more extensive investigations into hyperparameter configurations, such as the number of generated and retrieved documents will also be done.
\bibliographystyle{ACM-Reference-Format}
\bibliography{software}


\begin{thebibliography}{46}


\ifx \showCODEN    \undefined \def \showCODEN     #1{\unskip}     \fi
\ifx \showDOI      \undefined \def \showDOI       #1{#1}\fi
\ifx \showISBNx    \undefined \def \showISBNx     #1{\unskip}     \fi
\ifx \showISBNxiii \undefined \def \showISBNxiii  #1{\unskip}     \fi
\ifx \showISSN     \undefined \def \showISSN      #1{\unskip}     \fi
\ifx \showLCCN     \undefined \def \showLCCN      #1{\unskip}     \fi
\ifx \shownote     \undefined \def \shownote      #1{#1}          \fi
\ifx \showarticletitle \undefined \def \showarticletitle #1{#1}   \fi
\ifx \showURL      \undefined \def \showURL       {\relax}        \fi
\providecommand\bibfield[2]{#2}
\providecommand\bibinfo[2]{#2}
\providecommand\natexlab[1]{#1}
\providecommand\showeprint[2][]{arXiv:#2}

\bibitem[Askari et~al\mbox{.}(2023)]%
        {askari-etal-2023-expand}
\bibfield{author}{\bibinfo{person}{Arian Askari}, \bibinfo{person}{Mohammad Aliannejadi}, \bibinfo{person}{Chuan Meng}, \bibinfo{person}{Evangelos Kanoulas}, {and} \bibinfo{person}{Suzan Verberne}.} \bibinfo{year}{2023}\natexlab{}.
\newblock \showarticletitle{Expand, Highlight, Generate: {RL}-driven Document Generation for Passage Reranking}. In \bibinfo{booktitle}{\emph{Proceedings of the 2023 Conference on Empirical Methods in Natural Language Processing}}, \bibfield{editor}{\bibinfo{person}{Houda Bouamor}, \bibinfo{person}{Juan Pino}, {and} \bibinfo{person}{Kalika Bali}} (Eds.). \bibinfo{publisher}{Association for Computational Linguistics}, \bibinfo{address}{Singapore}, \bibinfo{pages}{10087--10099}.
\newblock
\urldef\tempurl%
\url{https://doi.org/10.18653/v1/2023.emnlp-main.623}
\showDOI{\tempurl}


\bibitem[Berant et~al\mbox{.}(2013)]%
        {berant2013semantic}
\bibfield{author}{\bibinfo{person}{Jonathan Berant}, \bibinfo{person}{Andrew Chou}, \bibinfo{person}{Roy Frostig}, {and} \bibinfo{person}{Percy Liang}.} \bibinfo{year}{2013}\natexlab{}.
\newblock \showarticletitle{Semantic parsing on freebase from question-answer pairs}. In \bibinfo{booktitle}{\emph{Proceedings of the 2013 conference on empirical methods in natural language processing}}. \bibinfo{pages}{1533--1544}.
\newblock


\bibitem[Brown et~al\mbox{.}(2020)]%
        {brown2020language}
\bibfield{author}{\bibinfo{person}{Tom Brown}, \bibinfo{person}{Benjamin Mann}, \bibinfo{person}{Nick Ryder}, \bibinfo{person}{Melanie Subbiah}, \bibinfo{person}{Jared~D Kaplan}, \bibinfo{person}{Prafulla Dhariwal}, \bibinfo{person}{Arvind Neelakantan}, \bibinfo{person}{Pranav Shyam}, \bibinfo{person}{Girish Sastry}, \bibinfo{person}{Amanda Askell}, {et~al\mbox{.}}} \bibinfo{year}{2020}\natexlab{}.
\newblock \showarticletitle{Language models are few-shot learners}.
\newblock \bibinfo{journal}{\emph{Advances in neural information processing systems}}  \bibinfo{volume}{33} (\bibinfo{year}{2020}), \bibinfo{pages}{1877--1901}.
\newblock


\bibitem[Chen et~al\mbox{.}(2020)]%
        {chen2020open}
\bibfield{author}{\bibinfo{person}{Wenhu Chen}, \bibinfo{person}{Ming-Wei Chang}, \bibinfo{person}{Eva Schlinger}, \bibinfo{person}{William Wang}, {and} \bibinfo{person}{William~W Cohen}.} \bibinfo{year}{2020}\natexlab{}.
\newblock \showarticletitle{Open question answering over tables and text}.
\newblock \bibinfo{journal}{\emph{arXiv preprint arXiv:2010.10439}} (\bibinfo{year}{2020}).
\newblock


\bibitem[Du et~al\mbox{.}(2022)]%
        {du2022glam}
\bibfield{author}{\bibinfo{person}{Nan Du}, \bibinfo{person}{Yanping Huang}, \bibinfo{person}{Andrew~M Dai}, \bibinfo{person}{Simon Tong}, \bibinfo{person}{Dmitry Lepikhin}, \bibinfo{person}{Yuanzhong Xu}, \bibinfo{person}{Maxim Krikun}, \bibinfo{person}{Yanqi Zhou}, \bibinfo{person}{Adams~Wei Yu}, \bibinfo{person}{Orhan Firat}, {et~al\mbox{.}}} \bibinfo{year}{2022}\natexlab{}.
\newblock \showarticletitle{Glam: Efficient scaling of language models with mixture-of-experts}. In \bibinfo{booktitle}{\emph{International Conference on Machine Learning}}. PMLR, \bibinfo{pages}{5547--5569}.
\newblock


\bibitem[Guu et~al\mbox{.}(2020)]%
        {guu2020retrieval}
\bibfield{author}{\bibinfo{person}{Kelvin Guu}, \bibinfo{person}{Kenton Lee}, \bibinfo{person}{Zora Tung}, \bibinfo{person}{Panupong Pasupat}, {and} \bibinfo{person}{Mingwei Chang}.} \bibinfo{year}{2020}\natexlab{}.
\newblock \showarticletitle{Retrieval augmented language model pre-training}. In \bibinfo{booktitle}{\emph{International conference on machine learning}}. PMLR, \bibinfo{pages}{3929--3938}.
\newblock


\bibitem[Han et~al\mbox{.}(2021)]%
        {han-etal-2021-exploring}
\bibfield{author}{\bibinfo{person}{Jiale Han}, \bibinfo{person}{Bo Cheng}, {and} \bibinfo{person}{Wei Lu}.} \bibinfo{year}{2021}\natexlab{}.
\newblock \showarticletitle{Exploring Task Difficulty for Few-Shot Relation Extraction}. In \bibinfo{booktitle}{\emph{Proceedings of the 2021 Conference on Empirical Methods in Natural Language Processing}}. \bibinfo{publisher}{Association for Computational Linguistics}, \bibinfo{address}{Online and Punta Cana, Dominican Republic}, \bibinfo{pages}{2605--2616}.
\newblock
\urldef\tempurl%
\url{https://doi.org/10.18653/v1/2021.emnlp-main.204}
\showDOI{\tempurl}


\bibitem[Hu et~al\mbox{.}(2021)]%
        {hu2021lora}
\bibfield{author}{\bibinfo{person}{Edward Hu}, \bibinfo{person}{Yelong Shen}, \bibinfo{person}{Phil Wallis}, \bibinfo{person}{Zeyuan Allen-Zhu}, \bibinfo{person}{Yuanzhi Li}, \bibinfo{person}{Lu Wang}, {and} \bibinfo{person}{Weizhu Chen}.} \bibinfo{year}{2021}\natexlab{}.
\newblock \bibinfo{title}{LoRA: Low-Rank Adaptation of Large Language Models}.
\newblock
\newblock
\showeprint[arxiv]{2106.09685}~[cs.CL]


\bibitem[Hunger(2005)]%
        {hunger2005floating}
\bibfield{author}{\bibinfo{person}{Raphael Hunger}.} \bibinfo{year}{2005}\natexlab{}.
\newblock \bibinfo{booktitle}{\emph{Floating point operations in matrix-vector calculus}}. Vol.~\bibinfo{volume}{2019}.
\newblock \bibinfo{publisher}{Munich University of Technology, Inst. for Circuit Theory and Signal~…}.
\newblock


\bibitem[Izacard and Grave(2020)]%
        {izacard2020leveraging}
\bibfield{author}{\bibinfo{person}{Gautier Izacard} {and} \bibinfo{person}{Edouard Grave}.} \bibinfo{year}{2020}\natexlab{}.
\newblock \showarticletitle{Leveraging passage retrieval with generative models for open domain question answering}.
\newblock \bibinfo{journal}{\emph{arXiv preprint arXiv:2007.01282}} (\bibinfo{year}{2020}).
\newblock


\bibitem[Joshi et~al\mbox{.}(2017)]%
        {joshi2017triviaqa}
\bibfield{author}{\bibinfo{person}{Mandar Joshi}, \bibinfo{person}{Eunsol Choi}, \bibinfo{person}{Daniel~S Weld}, {and} \bibinfo{person}{Luke Zettlemoyer}.} \bibinfo{year}{2017}\natexlab{}.
\newblock \showarticletitle{Triviaqa: A large scale distantly supervised challenge dataset for reading comprehension}.
\newblock \bibinfo{journal}{\emph{arXiv preprint arXiv:1705.03551}} (\bibinfo{year}{2017}).
\newblock


\bibitem[Kaplan et~al\mbox{.}(2020)]%
        {kaplan2020scaling}
\bibfield{author}{\bibinfo{person}{Jared Kaplan}, \bibinfo{person}{Sam McCandlish}, \bibinfo{person}{Tom Henighan}, \bibinfo{person}{Tom~B Brown}, \bibinfo{person}{Benjamin Chess}, \bibinfo{person}{Rewon Child}, \bibinfo{person}{Scott Gray}, \bibinfo{person}{Alec Radford}, \bibinfo{person}{Jeffrey Wu}, {and} \bibinfo{person}{Dario Amodei}.} \bibinfo{year}{2020}\natexlab{}.
\newblock \showarticletitle{Scaling laws for neural language models}.
\newblock \bibinfo{journal}{\emph{arXiv preprint arXiv:2001.08361}} (\bibinfo{year}{2020}).
\newblock


\bibitem[Karpukhin et~al\mbox{.}(2020)]%
        {karpukhin2020dense}
\bibfield{author}{\bibinfo{person}{Vladimir Karpukhin}, \bibinfo{person}{Barlas Oguz}, \bibinfo{person}{Sewon Min}, \bibinfo{person}{Patrick Lewis}, \bibinfo{person}{Ledell Wu}, \bibinfo{person}{Sergey Edunov}, \bibinfo{person}{Danqi Chen}, {and} \bibinfo{person}{Wen-tau Yih}.} \bibinfo{year}{2020}\natexlab{}.
\newblock \showarticletitle{Dense Passage Retrieval for Open-Domain Question Answering}. In \bibinfo{booktitle}{\emph{Proceedings of the 2020 Conference on Empirical Methods in Natural Language Processing (EMNLP)}}. \bibinfo{pages}{6769--6781}.
\newblock


\bibitem[Kwiatkowski et~al\mbox{.}(2019)]%
        {kwiatkowski2019natural}
\bibfield{author}{\bibinfo{person}{Tom Kwiatkowski}, \bibinfo{person}{Jennimaria Palomaki}, \bibinfo{person}{Olivia Redfield}, \bibinfo{person}{Michael Collins}, \bibinfo{person}{Ankur Parikh}, \bibinfo{person}{Chris Alberti}, \bibinfo{person}{Danielle Epstein}, \bibinfo{person}{Illia Polosukhin}, \bibinfo{person}{Jacob Devlin}, \bibinfo{person}{Kenton Lee}, {et~al\mbox{.}}} \bibinfo{year}{2019}\natexlab{}.
\newblock \showarticletitle{Natural questions: a benchmark for question answering research}.
\newblock \bibinfo{journal}{\emph{Transactions of the Association for Computational Linguistics}}  \bibinfo{volume}{7} (\bibinfo{year}{2019}), \bibinfo{pages}{453--466}.
\newblock


\bibitem[Lee et~al\mbox{.}(2020)]%
        {lee2020learning}
\bibfield{author}{\bibinfo{person}{Jinhyuk Lee}, \bibinfo{person}{Mujeen Sung}, \bibinfo{person}{Jaewoo Kang}, {and} \bibinfo{person}{Danqi Chen}.} \bibinfo{year}{2020}\natexlab{}.
\newblock \showarticletitle{Learning dense representations of phrases at scale}.
\newblock \bibinfo{journal}{\emph{arXiv preprint arXiv:2012.12624}} (\bibinfo{year}{2020}).
\newblock


\bibitem[Lee et~al\mbox{.}(2021)]%
        {lee2021learning}
\bibfield{author}{\bibinfo{person}{Jinhyuk Lee}, \bibinfo{person}{Mujeen Sung}, \bibinfo{person}{Jaewoo Kang}, {and} \bibinfo{person}{Danqi Chen}.} \bibinfo{year}{2021}\natexlab{}.
\newblock \bibinfo{title}{Learning Dense Representations of Phrases at Scale}.
\newblock
\newblock
\showeprint[arxiv]{2012.12624}~[cs.CL]


\bibitem[Lee et~al\mbox{.}(2019)]%
        {lee2019latent}
\bibfield{author}{\bibinfo{person}{Kenton Lee}, \bibinfo{person}{Ming-Wei Chang}, {and} \bibinfo{person}{Kristina Toutanova}.} \bibinfo{year}{2019}\natexlab{}.
\newblock \showarticletitle{Latent retrieval for weakly supervised open domain question answering}.
\newblock \bibinfo{journal}{\emph{arXiv preprint arXiv:1906.00300}} (\bibinfo{year}{2019}).
\newblock


\bibitem[Levine et~al\mbox{.}(2022)]%
        {levine2022standing}
\bibfield{author}{\bibinfo{person}{Yoav Levine}, \bibinfo{person}{Itay Dalmedigos}, \bibinfo{person}{Ori Ram}, \bibinfo{person}{Yoel Zeldes}, \bibinfo{person}{Daniel Jannai}, \bibinfo{person}{Dor Muhlgay}, \bibinfo{person}{Yoni Osin}, \bibinfo{person}{Opher Lieber}, \bibinfo{person}{Barak Lenz}, \bibinfo{person}{Shai Shalev-Shwartz}, {et~al\mbox{.}}} \bibinfo{year}{2022}\natexlab{}.
\newblock \showarticletitle{Standing on the shoulders of giant frozen language models}.
\newblock \bibinfo{journal}{\emph{arXiv preprint arXiv:2204.10019}} (\bibinfo{year}{2022}).
\newblock


\bibitem[Lewis et~al\mbox{.}(2020)]%
        {lewis2020retrieval}
\bibfield{author}{\bibinfo{person}{Patrick Lewis}, \bibinfo{person}{Ethan Perez}, \bibinfo{person}{Aleksandra Piktus}, \bibinfo{person}{Fabio Petroni}, \bibinfo{person}{Vladimir Karpukhin}, \bibinfo{person}{Naman Goyal}, \bibinfo{person}{Heinrich K{\"u}ttler}, \bibinfo{person}{Mike Lewis}, \bibinfo{person}{Wen-tau Yih}, \bibinfo{person}{Tim Rockt{\"a}schel}, {et~al\mbox{.}}} \bibinfo{year}{2020}\natexlab{}.
\newblock \showarticletitle{Retrieval-augmented generation for knowledge-intensive nlp tasks}.
\newblock \bibinfo{journal}{\emph{Advances in Neural Information Processing Systems}}  \bibinfo{volume}{33} (\bibinfo{year}{2020}), \bibinfo{pages}{9459--9474}.
\newblock


\bibitem[Liu(2022)]%
        {Liu_LlamaIndex_2022}
\bibfield{author}{\bibinfo{person}{Jerry Liu}.} \bibinfo{year}{2022}\natexlab{}.
\newblock \bibinfo{booktitle}{\emph{{LlamaIndex}}}.
\newblock
\urldef\tempurl%
\url{https://doi.org/10.5281/zenodo.1234}
\showDOI{\tempurl}


\bibitem[Liu et~al\mbox{.}(2018)]%
        {liu2018learning}
\bibfield{author}{\bibinfo{person}{Yanbin Liu}, \bibinfo{person}{Juho Lee}, \bibinfo{person}{Minseop Park}, \bibinfo{person}{Saehoon Kim}, \bibinfo{person}{Eunho Yang}, \bibinfo{person}{Sung~Ju Hwang}, {and} \bibinfo{person}{Yi Yang}.} \bibinfo{year}{2018}\natexlab{}.
\newblock \showarticletitle{Learning to propagate labels: Transductive propagation network for few-shot learning}.
\newblock \bibinfo{journal}{\emph{arXiv preprint arXiv:1805.10002}} (\bibinfo{year}{2018}).
\newblock


\bibitem[Ni et~al\mbox{.}(2021)]%
        {ni2021large}
\bibfield{author}{\bibinfo{person}{Jianmo Ni}, \bibinfo{person}{Chen Qu}, \bibinfo{person}{Jing Lu}, \bibinfo{person}{Zhuyun Dai}, \bibinfo{person}{Gustavo~Hern{\'a}ndez {\'A}brego}, \bibinfo{person}{Ji Ma}, \bibinfo{person}{Vincent~Y Zhao}, \bibinfo{person}{Yi Luan}, \bibinfo{person}{Keith~B Hall}, \bibinfo{person}{Ming-Wei Chang}, {et~al\mbox{.}}} \bibinfo{year}{2021}\natexlab{}.
\newblock \showarticletitle{Large dual encoders are generalizable retrievers}.
\newblock \bibinfo{journal}{\emph{arXiv preprint arXiv:2112.07899}} (\bibinfo{year}{2021}).
\newblock


\bibitem[Oguz et~al\mbox{.}(2020)]%
        {oguz2020unik}
\bibfield{author}{\bibinfo{person}{Barlas Oguz}, \bibinfo{person}{Xilun Chen}, \bibinfo{person}{Vladimir Karpukhin}, \bibinfo{person}{Stan Peshterliev}, \bibinfo{person}{Dmytro Okhonko}, \bibinfo{person}{Michael Schlichtkrull}, \bibinfo{person}{Sonal Gupta}, \bibinfo{person}{Yashar Mehdad}, {and} \bibinfo{person}{Scott Yih}.} \bibinfo{year}{2020}\natexlab{}.
\newblock \showarticletitle{Unik-qa: Unified representations of structured and unstructured knowledge for open-domain question answering}.
\newblock \bibinfo{journal}{\emph{arXiv preprint arXiv:2012.14610}} (\bibinfo{year}{2020}).
\newblock


\bibitem[Ouyang et~al\mbox{.}(2022)]%
        {ouyang2022training}
\bibfield{author}{\bibinfo{person}{Long Ouyang}, \bibinfo{person}{Jeffrey Wu}, \bibinfo{person}{Xu Jiang}, \bibinfo{person}{Diogo Almeida}, \bibinfo{person}{Carroll Wainwright}, \bibinfo{person}{Pamela Mishkin}, \bibinfo{person}{Chong Zhang}, \bibinfo{person}{Sandhini Agarwal}, \bibinfo{person}{Katarina Slama}, \bibinfo{person}{Alex Ray}, {et~al\mbox{.}}} \bibinfo{year}{2022}\natexlab{}.
\newblock \showarticletitle{Training language models to follow instructions with human feedback}.
\newblock \bibinfo{journal}{\emph{Advances in Neural Information Processing Systems}}  \bibinfo{volume}{35} (\bibinfo{year}{2022}), \bibinfo{pages}{27730--27744}.
\newblock


\bibitem[Paszke et~al\mbox{.}(2019)]%
        {paszke2019pytorch}
\bibfield{author}{\bibinfo{person}{Adam Paszke}, \bibinfo{person}{Sam Gross}, \bibinfo{person}{Francisco Massa}, \bibinfo{person}{Adam Lerer}, \bibinfo{person}{James Bradbury}, \bibinfo{person}{Gregory Chanan}, \bibinfo{person}{Trevor Killeen}, \bibinfo{person}{Zeming Lin}, \bibinfo{person}{Natalia Gimelshein}, \bibinfo{person}{Luca Antiga}, {et~al\mbox{.}}} \bibinfo{year}{2019}\natexlab{}.
\newblock \showarticletitle{Pytorch: An imperative style, high-performance deep learning library}.
\newblock \bibinfo{journal}{\emph{Advances in neural information processing systems}}  \bibinfo{volume}{32} (\bibinfo{year}{2019}).
\newblock


\bibitem[Qaiser and Ali(2018)]%
        {article_tiidf}
\bibfield{author}{\bibinfo{person}{Shahzad Qaiser} {and} \bibinfo{person}{Ramsha Ali}.} \bibinfo{year}{2018}\natexlab{}.
\newblock \showarticletitle{Text Mining: Use of TF-IDF to Examine the Relevance of Words to Documents}.
\newblock \bibinfo{journal}{\emph{International Journal of Computer Applications}}  \bibinfo{volume}{181} (\bibinfo{date}{07} \bibinfo{year}{2018}).
\newblock
\urldef\tempurl%
\url{https://doi.org/10.5120/ijca2018917395}
\showDOI{\tempurl}


\bibitem[Radford et~al\mbox{.}(2019)]%
        {radford2019language}
\bibfield{author}{\bibinfo{person}{Alec Radford}, \bibinfo{person}{Jeffrey Wu}, \bibinfo{person}{Rewon Child}, \bibinfo{person}{David Luan}, \bibinfo{person}{Dario Amodei}, \bibinfo{person}{Ilya Sutskever}, {et~al\mbox{.}}} \bibinfo{year}{2019}\natexlab{}.
\newblock \showarticletitle{Language models are unsupervised multitask learners}.
\newblock \bibinfo{journal}{\emph{OpenAI blog}} \bibinfo{volume}{1}, \bibinfo{number}{8} (\bibinfo{year}{2019}), \bibinfo{pages}{9}.
\newblock


\bibitem[Rae et~al\mbox{.}(2021)]%
        {rae2021scaling}
\bibfield{author}{\bibinfo{person}{Jack~W Rae}, \bibinfo{person}{Sebastian Borgeaud}, \bibinfo{person}{Trevor Cai}, \bibinfo{person}{Katie Millican}, \bibinfo{person}{Jordan Hoffmann}, \bibinfo{person}{Francis Song}, \bibinfo{person}{John Aslanides}, \bibinfo{person}{Sarah Henderson}, \bibinfo{person}{Roman Ring}, \bibinfo{person}{Susannah Young}, {et~al\mbox{.}}} \bibinfo{year}{2021}\natexlab{}.
\newblock \showarticletitle{Scaling language models: Methods, analysis \& insights from training gopher}.
\newblock \bibinfo{journal}{\emph{arXiv preprint arXiv:2112.11446}} (\bibinfo{year}{2021}).
\newblock


\bibitem[Raffel et~al\mbox{.}(2020)]%
        {raffel2020exploring}
\bibfield{author}{\bibinfo{person}{Colin Raffel}, \bibinfo{person}{Noam Shazeer}, \bibinfo{person}{Adam Roberts}, \bibinfo{person}{Katherine Lee}, \bibinfo{person}{Sharan Narang}, \bibinfo{person}{Michael Matena}, \bibinfo{person}{Yanqi Zhou}, \bibinfo{person}{Wei Li}, {and} \bibinfo{person}{Peter~J Liu}.} \bibinfo{year}{2020}\natexlab{}.
\newblock \showarticletitle{Exploring the limits of transfer learning with a unified text-to-text transformer}.
\newblock \bibinfo{journal}{\emph{The Journal of Machine Learning Research}} \bibinfo{volume}{21}, \bibinfo{number}{1} (\bibinfo{year}{2020}), \bibinfo{pages}{5485--5551}.
\newblock


\bibitem[Rajbhandari et~al\mbox{.}(2020)]%
        {deepspeed_zero}
\bibfield{author}{\bibinfo{person}{Samyam Rajbhandari}, \bibinfo{person}{Jeff Rasley}, \bibinfo{person}{Olatunji Ruwase}, {and} \bibinfo{person}{Yuxiong He}.} \bibinfo{year}{2020}\natexlab{}.
\newblock \showarticletitle{ZeRO: Memory Optimizations toward Training Trillion Parameter Models}. In \bibinfo{booktitle}{\emph{Proceedings of the International Conference for High Performance Computing, Networking, Storage and Analysis}} (Atlanta, Georgia) \emph{(\bibinfo{series}{SC '20})}. \bibinfo{publisher}{IEEE Press}, Article \bibinfo{articleno}{20}, \bibinfo{numpages}{16}~pages.
\newblock
\showISBNx{9781728199986}


\bibitem[Reimers and Gurevych(2019)]%
        {reimers2019sentencebert}
\bibfield{author}{\bibinfo{person}{Nils Reimers} {and} \bibinfo{person}{Iryna Gurevych}.} \bibinfo{year}{2019}\natexlab{}.
\newblock \bibinfo{title}{Sentence-BERT: Sentence Embeddings using Siamese BERT-Networks}.
\newblock
\newblock
\showeprint[arxiv]{1908.10084}~[cs.CL]


\bibitem[Roberts et~al\mbox{.}(2020)]%
        {roberts2020much}
\bibfield{author}{\bibinfo{person}{Adam Roberts}, \bibinfo{person}{Colin Raffel}, {and} \bibinfo{person}{Noam Shazeer}.} \bibinfo{year}{2020}\natexlab{}.
\newblock \showarticletitle{How much knowledge can you pack into the parameters of a language model?}
\newblock \bibinfo{journal}{\emph{arXiv preprint arXiv:2002.08910}} (\bibinfo{year}{2020}).
\newblock


\bibitem[Rosa et~al\mbox{.}(2021)]%
        {rosa2021yes}
\bibfield{author}{\bibinfo{person}{Guilherme~Moraes Rosa}, \bibinfo{person}{Ruan~Chaves Rodrigues}, \bibinfo{person}{Roberto Lotufo}, {and} \bibinfo{person}{Rodrigo Nogueira}.} \bibinfo{year}{2021}\natexlab{}.
\newblock \bibinfo{title}{Yes, BM25 is a Strong Baseline for Legal Case Retrieval}.
\newblock
\newblock
\showeprint[arxiv]{2105.05686}~[cs.IR]


\bibitem[Sachan et~al\mbox{.}(2022)]%
        {sachan2022improving}
\bibfield{author}{\bibinfo{person}{Devendra~Singh Sachan}, \bibinfo{person}{Mike Lewis}, \bibinfo{person}{Mandar Joshi}, \bibinfo{person}{Armen Aghajanyan}, \bibinfo{person}{Wen-tau Yih}, \bibinfo{person}{Joelle Pineau}, {and} \bibinfo{person}{Luke Zettlemoyer}.} \bibinfo{year}{2022}\natexlab{}.
\newblock \showarticletitle{Improving Passage Retrieval with Zero-Shot Question Generation}.
\newblock  (\bibinfo{year}{2022}).
\newblock
\urldef\tempurl%
\url{https://arxiv.org/abs/2204.07496}
\showURL{%
\tempurl}


\bibitem[Seo et~al\mbox{.}(2018)]%
        {seo2018bidirectional}
\bibfield{author}{\bibinfo{person}{Minjoon Seo}, \bibinfo{person}{Aniruddha Kembhavi}, \bibinfo{person}{Ali Farhadi}, {and} \bibinfo{person}{Hannaneh Hajishirzi}.} \bibinfo{year}{2018}\natexlab{}.
\newblock \bibinfo{title}{Bidirectional Attention Flow for Machine Comprehension}.
\newblock
\newblock
\showeprint[arxiv]{1611.01603}~[cs.CL]


\bibitem[Seo et~al\mbox{.}(2019)]%
        {seo2019real}
\bibfield{author}{\bibinfo{person}{Minjoon Seo}, \bibinfo{person}{Jinhyuk Lee}, \bibinfo{person}{Tom Kwiatkowski}, \bibinfo{person}{Ankur~P Parikh}, \bibinfo{person}{Ali Farhadi}, {and} \bibinfo{person}{Hannaneh Hajishirzi}.} \bibinfo{year}{2019}\natexlab{}.
\newblock \showarticletitle{Real-time open-domain question answering with dense-sparse phrase index}.
\newblock \bibinfo{journal}{\emph{arXiv preprint arXiv:1906.05807}} (\bibinfo{year}{2019}).
\newblock


\bibitem[Singh et~al\mbox{.}(2021)]%
        {singh2021end}
\bibfield{author}{\bibinfo{person}{Devendra Singh}, \bibinfo{person}{Siva Reddy}, \bibinfo{person}{Will Hamilton}, \bibinfo{person}{Chris Dyer}, {and} \bibinfo{person}{Dani Yogatama}.} \bibinfo{year}{2021}\natexlab{}.
\newblock \showarticletitle{End-to-end training of multi-document reader and retriever for open-domain question answering}.
\newblock \bibinfo{journal}{\emph{Advances in Neural Information Processing Systems}}  \bibinfo{volume}{34} (\bibinfo{year}{2021}), \bibinfo{pages}{25968--25981}.
\newblock


\bibitem[Touvron et~al\mbox{.}(2023)]%
        {touvron2023llama}
\bibfield{author}{\bibinfo{person}{Hugo Touvron}, \bibinfo{person}{Thibaut Lavril}, \bibinfo{person}{Gautier Izacard}, \bibinfo{person}{Xavier Martinet}, \bibinfo{person}{Marie-Anne Lachaux}, \bibinfo{person}{Timoth{\'e}e Lacroix}, \bibinfo{person}{Baptiste Rozi{\`e}re}, \bibinfo{person}{Naman Goyal}, \bibinfo{person}{Eric Hambro}, \bibinfo{person}{Faisal Azhar}, {et~al\mbox{.}}} \bibinfo{year}{2023}\natexlab{}.
\newblock \showarticletitle{Llama: Open and efficient foundation language models}.
\newblock \bibinfo{journal}{\emph{arXiv preprint arXiv:2302.13971}} (\bibinfo{year}{2023}).
\newblock


\bibitem[Vaswani et~al\mbox{.}(2017)]%
        {vaswani2017attention}
\bibfield{author}{\bibinfo{person}{Ashish Vaswani}, \bibinfo{person}{Noam Shazeer}, \bibinfo{person}{Niki Parmar}, \bibinfo{person}{Jakob Uszkoreit}, \bibinfo{person}{Llion Jones}, \bibinfo{person}{Aidan~N Gomez}, \bibinfo{person}{{\L}ukasz Kaiser}, {and} \bibinfo{person}{Illia Polosukhin}.} \bibinfo{year}{2017}\natexlab{}.
\newblock \showarticletitle{Attention Is All You Need}. In \bibinfo{booktitle}{\emph{Advances in Neural Information Processing Systems}}.
\newblock
\urldef\tempurl%
\url{https://proceedings.neurips.cc/paper/2017/hash/3f5ee243547dee91fbd053c1c4a845aa-Abstract.html}
\showURL{%
\tempurl}


\bibitem[Wang et~al\mbox{.}(2023)]%
        {wang2023rfid}
\bibfield{author}{\bibinfo{person}{Cunxiang Wang}, \bibinfo{person}{Haofei Yu}, {and} \bibinfo{person}{Yue Zhang}.} \bibinfo{year}{2023}\natexlab{}.
\newblock \showarticletitle{RFiD: Towards Rational Fusion-in-Decoder for Open-Domain Question Answering}.
\newblock \bibinfo{journal}{\emph{arXiv preprint arXiv:2305.17041}} (\bibinfo{year}{2023}).
\newblock


\bibitem[Wei et~al\mbox{.}(2021)]%
        {wei2021finetuned}
\bibfield{author}{\bibinfo{person}{Jason Wei}, \bibinfo{person}{Maarten Bosma}, \bibinfo{person}{Vincent~Y Zhao}, \bibinfo{person}{Kelvin Guu}, \bibinfo{person}{Adams~Wei Yu}, \bibinfo{person}{Brian Lester}, \bibinfo{person}{Nan Du}, \bibinfo{person}{Andrew~M Dai}, {and} \bibinfo{person}{Quoc~V Le}.} \bibinfo{year}{2021}\natexlab{}.
\newblock \showarticletitle{Finetuned language models are zero-shot learners}.
\newblock \bibinfo{journal}{\emph{arXiv preprint arXiv:2109.01652}} (\bibinfo{year}{2021}).
\newblock


\bibitem[Xiao et~al\mbox{.}(2022)]%
        {xiao2022distill}
\bibfield{author}{\bibinfo{person}{Shitao Xiao}, \bibinfo{person}{Zheng Liu}, \bibinfo{person}{Weihao Han}, \bibinfo{person}{Jianjin Zhang}, \bibinfo{person}{Defu Lian}, \bibinfo{person}{Yeyun Gong}, \bibinfo{person}{Qi Chen}, \bibinfo{person}{Fan Yang}, \bibinfo{person}{Hao Sun}, \bibinfo{person}{Yingxia Shao}, {et~al\mbox{.}}} \bibinfo{year}{2022}\natexlab{}.
\newblock \showarticletitle{Distill-vq: Learning retrieval oriented vector quantization by distilling knowledge from dense embeddings}. In \bibinfo{booktitle}{\emph{Proceedings of the 45th International ACM SIGIR Conference on Research and Development in Information Retrieval}}. \bibinfo{pages}{1513--1523}.
\newblock


\bibitem[Yang et~al\mbox{.}(2019)]%
        {yang2019end}
\bibfield{author}{\bibinfo{person}{Wei Yang}, \bibinfo{person}{Yuqing Xie}, \bibinfo{person}{Aileen Lin}, \bibinfo{person}{Xingyu Li}, \bibinfo{person}{Luchen Tan}, \bibinfo{person}{Kun Xiong}, \bibinfo{person}{Ming Li}, {and} \bibinfo{person}{Jimmy Lin}.} \bibinfo{year}{2019}\natexlab{}.
\newblock \showarticletitle{End-to-end open-domain question answering with bertserini}.
\newblock \bibinfo{journal}{\emph{arXiv preprint arXiv:1902.01718}} (\bibinfo{year}{2019}).
\newblock


\bibitem[Yu et~al\mbox{.}(2022)]%
        {yu2022generate}
\bibfield{author}{\bibinfo{person}{Wenhao Yu}, \bibinfo{person}{Dan Iter}, \bibinfo{person}{Shuohang Wang}, \bibinfo{person}{Yichong Xu}, \bibinfo{person}{Mingxuan Ju}, \bibinfo{person}{Soumya Sanyal}, \bibinfo{person}{Chenguang Zhu}, \bibinfo{person}{Michael Zeng}, {and} \bibinfo{person}{Meng Jiang}.} \bibinfo{year}{2022}\natexlab{}.
\newblock \showarticletitle{Generate rather than retrieve: Large language models are strong context generators}.
\newblock \bibinfo{journal}{\emph{arXiv preprint arXiv:2209.10063}} (\bibinfo{year}{2022}).
\newblock


\bibitem[Zhang et~al\mbox{.}(2023)]%
        {zhang2023merging}
\bibfield{author}{\bibinfo{person}{Yunxiang Zhang}, \bibinfo{person}{Muhammad Khalifa}, \bibinfo{person}{Lajanugen Logeswaran}, \bibinfo{person}{Moontae Lee}, \bibinfo{person}{Honglak Lee}, {and} \bibinfo{person}{Lu Wang}.} \bibinfo{year}{2023}\natexlab{}.
\newblock \showarticletitle{Merging Generated and Retrieved Knowledge for Open-Domain QA}.
\newblock \bibinfo{journal}{\emph{arXiv preprint arXiv:2310.14393}} (\bibinfo{year}{2023}).
\newblock


\bibitem[Zhu et~al\mbox{.}(2021)]%
        {zhu2021retrieving}
\bibfield{author}{\bibinfo{person}{Fengbin Zhu}, \bibinfo{person}{Wenqiang Lei}, \bibinfo{person}{Chao Wang}, \bibinfo{person}{Jianming Zheng}, \bibinfo{person}{Soujanya Poria}, {and} \bibinfo{person}{Tat-Seng Chua}.} \bibinfo{year}{2021}\natexlab{}.
\newblock \showarticletitle{Retrieving and reading: A comprehensive survey on open-domain question answering}.
\newblock \bibinfo{journal}{\emph{arXiv preprint arXiv:2101.00774}} (\bibinfo{year}{2021}).
\newblock


\end{thebibliography}

\appendix
\section{APPENDIX}
\begin{table}[h]
\caption{ Comparison of Generated Answers for Temporal Questions in the TQA Dataset}
\begin{minipage}{\linewidth}
\small
\centering
\begin{tabularx}{\linewidth}{@{} X X X X X @{}}
\toprule
\textbf{Questions from TQA test} & \textbf{TQA Label} & \textbf{GRG} & \textbf{Google} & \textbf{GPT} \\
\midrule
 What star sign is Jamie Lee Curtis? & Scorpio & Scorpio & Sagittarius & Sagittarius  \\
\\
\midrule
Which Lloyd Webber musical premiered in the US on 10th December 1993? & Sunset Blvd & Sunset Boulevard & Sunset Boulevard & Sunset Boulevard  \\
\midrule
Which actress was voted Miss Greenwich Village in 1942? & Lauren Becal & Joanne Woodward & Lauren Bacall & No answer \\

\bottomrule
\end{tabularx}
\label{table:examples1}
\end{minipage}
\end{table}
\subsection{Computational Cost Analysis}
\label{cost_anaysis}
In this section, we compare the computational costs of using Dense Passage Retrieval (DPR) and InstructGPT for document retrieval and generation, respectively. We use DPR implemented with the T5 model~\citep{roberts2020much}, which has approximately 220 million parameters, and InstructGPT in its largest configuration with 175 billion parameters.

\subsubsection{Cost Metrics}

We estimate the computational cost in terms of Floating Point Operations (FLOPs) per token, a metric introduced by~\citet{kaplan2020scaling} and~\citet{hunger2005floating}. FLOPs provide an estimate of the computational cost required by a model to process a given input. However, it’s important to note that FLOPs are not a direct measure of real-world computing costs, as factors such as latency, power consumption, and hardware efficiency can vary widely.
\subsubsection{Cost Comparison}
The computational costs of DPR and InstructGPT are compared as follow $DPR = 
 O(|q| \times |D|)$ and  $InstructGPT=O(|q| \times |T|)$ . Here, $|q|$ represents the length of the query, $|D|$ is the number of documents in the corpus, and $|T|$ is the number of tokens in the document.

As shown in Table~\ref{cost_dpr_instruct}, the cost of using DPR is proportional to the number of documents in the corpus, while the cost of using InstructGPT is proportional to the number of tokens in the document. This implies that InstructGPT is more efficient for generating documents, while DPR is more efficient for retrieving documents.

\subsubsection{Model-Specific Costs}
We further break down the costs for each model:
\begin{enumerate}
    \item Document Retriever: T5 Model - The T5 model, with approximately 220 million parameters and a token limit of 512 tokens per document and question, is used for document retrieval. The computational costs for encoding all 21 million Wikipedia documents and retrieving documents for a given question using T5 are calculated as follows:
\begin{align*}
\text{FLOPs} &= 220 \times 10^6 \times 21 \times 10^6 \times 512 \\
&= 2.84 \times 10^{18} \text{ FLOPs}
\end{align*}

\begin{align*}
\text{FLOPs} &= 220 \times 10^6 \times 20 \\
&\quad+ 21 \times 10^6 \times (768 + 768 - 1) \\
&= 3.77 \times 10^{11} \text{ FLOPs}
\end{align*}
\item Document Generator: InstructGPT Model - The InstructGPT model, with 175 billion parameters and a token limit of 512 tokens per document, is used for document generation. The computational cost for generating 10 documents for a given question with 100 words each using InstructGPT is calculated as follows:
\begin{align*}
\text{FLOPs} &= 175 \times 10^9 \times 10 \times 100 = 1.75 \times 10^{14} \text{ FLOPs}
\end{align*}

\item Document Generator Retriever - The cost of encoding all 10 documents with 100 words each using the T5 model is calculated as follows:

\begin{align*}
\text{FLOPs} &= 220 \times 10^6 \times 10 \times 100 = 2.2 \times 10^{12} \text{ FLOPs}
\end{align*}

\item The LLAMA model, with 7 billion parameters and a token limit of 512 tokens per document, is used for retrieving documents. The computational cost for retrieving 5 documents using LLAMA is calculated as follows:
\begin{align*}
\text{FLOPs} &= 7 \times 10^9 \times 5 \times 128 = 4.48 \times 10^{12} \text{ FLOPs}
\end{align*}
\end{enumerate}

\begin{table}[h]
\centering
\caption{NQ Dataset Comparison}
\small
\begin{tabularx}{\columnwidth}{@{} X X X X @{}}
\toprule
\textbf{Question} & \textbf{True Answer} & \textbf{GRG} & \textbf{GenRead} \\
\midrule

Who got the first Nobel Prize in Physics? & Wilhelm Conrad Röntgen & Wilhelm Conrad Röntgen & Wilhelm Conrad Röntgen\\

\midrule
When was coffee first made into a drink? & 15th century & 15th century & the 10th century\\
\midrule
Who won the MVP for the national league? & Stanton, Giancarlo  & Giancarlo Stanton & Christian Yelich\\
\midrule
Where do the greasers live in the outsiders? & Tulsa, Oklahoma & Tulsa, Oklahoma & Oklahoma\\
\midrule
Who played lionel in "As time goes by"?  &  Geoffrey Dyson Palmer, OBE  & Geoffrey Dyson Palmer & Geoffrey Palmer\\
\bottomrule
\end{tabularx}

\label{table:examples}
\end{table}

\subsection{Case study}

\subsubsection{NQ Case Study}
Our model's performance was evaluated using the NQ and TQA datasets, comparing GRG with GenRead. Table~\ref{table:examples} illustrates GRG's ability to generate accurate answers that align with the true answers for example questions taken from NQ dataset, demonstrating an improved understanding of the questions.

\subsubsection{TQA Case Study}
Further analysis with the TQA dataset highlights GRG's robustness, providing accurate answers and considering multiple valid responses. Table~\ref{table:examples1} shows GRG's performance in comparison to Google and GPT.

\end{document}